# Learning from Pseudo-Randomness with an Artificial Neural Network – Does God Play Pseudo-Dice?

Fenglei Fan, Ge Wang[1]
Biomedical Imaging Center, BME/CBIS
Rensselaer Polytechnic Institute, Troy, New York, USA, 12180

*Abstract* — **Inspired by the fact that the neural network, as the mainstream method for machine learning, has brought successes in many application areas, here we propose to use this approach for decoding hidden correlation among pseudo-random data and predicting events accordingly. With a simple neural network structure and a typical training procedure, we demonstrate the learning and prediction power of the neural network in pseudo-random environments. Finally, we postulate that the high sensitivity and efficiency of the neural network may allow to learn on a low-dimensional manifold in a high-dimensional space of pseudo-random events and critically test if there could be any fundamental difference between quantum randomness and pseudo randomness, which is equivalent to the classic question: Does God play dice?**

## I. INTRODUCTION

In the field of machine learning, neural networks, especially convolutional neural networks (CNNs) have recently delivered surprising outcomes in many types of applications such as classification, prediction [1-2], analysis [3-4], and image reconstruction [5-6]. Encouraged by the huge potential of neural networks, great efforts are being made to improve the network performance [7] and [8] and find new applications.

Machine learning can be performed in different modes. Supervised learning with a neural network involves the followed two steps: given a label y to every input sample x, forming a training dataset in the format of pairs $(x_1, y_1),(x_2, y_2)...(x_m, y_m)$; and then the neural network is trained using the backpropagation algorithm [9] to optimize the network function from inputs to outputs. Usually, the labels assigned to the inputs are well defined, which means that the output is unique to reflect a deterministic input-output relationship. However, this view is invalid in probabilistic scenarios. Statistical learning can also be performed using the neural network approach [10]; for example, to find high likelihood solutions while filtering out random noise. In this paper, we focus on how much a neural network could learn from pseudo-random noise, which is an extreme case of statistical learning.

As an initial investigation along this direction, we will use a popular multilayer neural network to predict the "next number" in a 0-1 pseudo-random number sequence. Pseudo-random numbers are closely related to mathematical chaotic behaviors [11-12]. Although chaotic time series prediction was investigated using machine learning methods [13-16], few peer-reviewed studies [17] were reported on pseudo-random number prediction, which were neither systematically designed nor rigorously tested.

In this paper, we provide a representative (0-1 random numbers), systematic (big datasets) and definitive (in the sense of 5σ) analysis on this topic. In the next section, we describe our experimental designs and key results. In the third, we discuss improvements and implications of our findings, especially how to test if quantum noise is really random or just pseudo-random.

## II. NUMERICAL STUDIES AND RESULTS

### A. *Learning from a Single Pseudo-Random Sequence*

$\pi$ is one of the most famous transcendental numbers, widely used in almost every discipline. However, there remain interesting questions on this number. For example, is an infinite sequence of its digits a realization of a perfect random mechanism or not? In other words, could the presence of a given digit be, to some definite degree, predicted based on a sufficiently large number of digits in a neighborhood around the digit of interest?

In our study, we cast this exemplary question into a binary version, which is even more challenging. Specifically, we binarized the sequence of $\pi$ with the threshold 5 to obtain a 0-1 sequence $\pi'$, as shown in Figure 1. Then, we used an artificial neural network with the configuration of 6-30-20-1 to predict the seventh digit from its precedent consecutive six numbers. The training dataset consisted of 40,000 instances made from the 1st to 40,007th digits of $\pi'$. Then, the trained network was evaluated by the two testing datasets: T1 with 900,000 instances from the 100,000th to 1,000,006th digits, and T2 with 9,000,000 instances from the 999,000th to 9,999,006th digits.

[1] Corresponding Author: G. Wang (wangg6@rpi.edu, 518-698-2500)

Furthermore, each testing dataset was divided into 9 subsets for repeated tests.

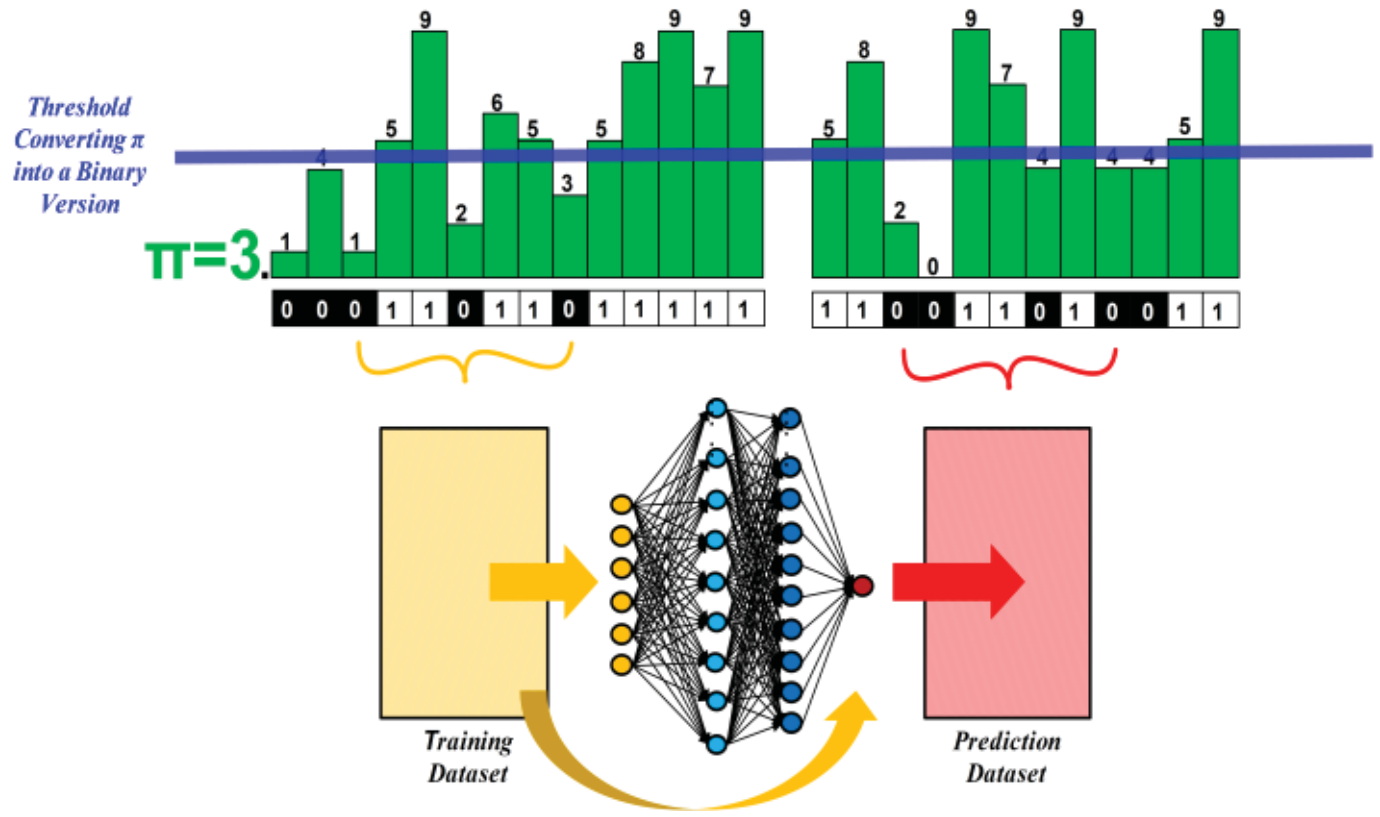

**Figure 1.** Learning from the pseudo-randomness of $\pi$. After the sequence $\pi$ is binarized into the black-white sequence $\pi'$, a typical feedforward neural network was trained to predict the next digit based on the latest six digits, and tested for statistical significance.

In the learning processing, the Levenberg-Marquart backpropagation algorithm was utilized. The weights and biases of the neural network were randomly initialized using the "*initnw*" function of Matlab. The learning parameters were empirically set to momentum factor=0.95, learning rate=0.05, and maximum epoch=40. The performance was defined by the number of successful predictions divided by the total number of instances.

First, we trained the network to predict T1, with the trained network being referred to as "Net 1". For the training set, 51.28% of the total instances were successfully predicted. For the 9 subgroups in T1, the correct prediction rates were found to be 0.50120, 0.50063, 0.50194, 0.50141, 0.49987, 0.50083, 0.50128, 0.50138, and 0.50189 respectively. Then, we independently retrained the network to predict T2, with the new network "Net 2". In the run, the training performance was the correct prediction rate 51.03%. For the 9 subgroups in T2, the correct prediction rates were 0.500344, 0.500465, 0.499750, 0.500118, 0.500465, 0.500239, 0.500326, 0.500692, and 0.500167 respectively. These results are encouraging, since our network is rather simplistic, and the probability of 0 or 1 are theoretically equal [18].

In the viewpoint of probability, the binary prediction of 0 or 1 in the sequence $\pi'$ will be totally random if

*Probability (Next Digit =1 | Other Digits) = Probability (Next Digit =0 | Other Digits) = 0.5* (1)

Eq. (1) means that any prediction of the next digit will have equal chance of being correct or incorrect. However, if Eq. (1) does not hold, then what the next digit will be somehow correlated to other digits in the sequence $\pi'$. Our above-described data shows that the latter is the case for the sequence $\pi'$.

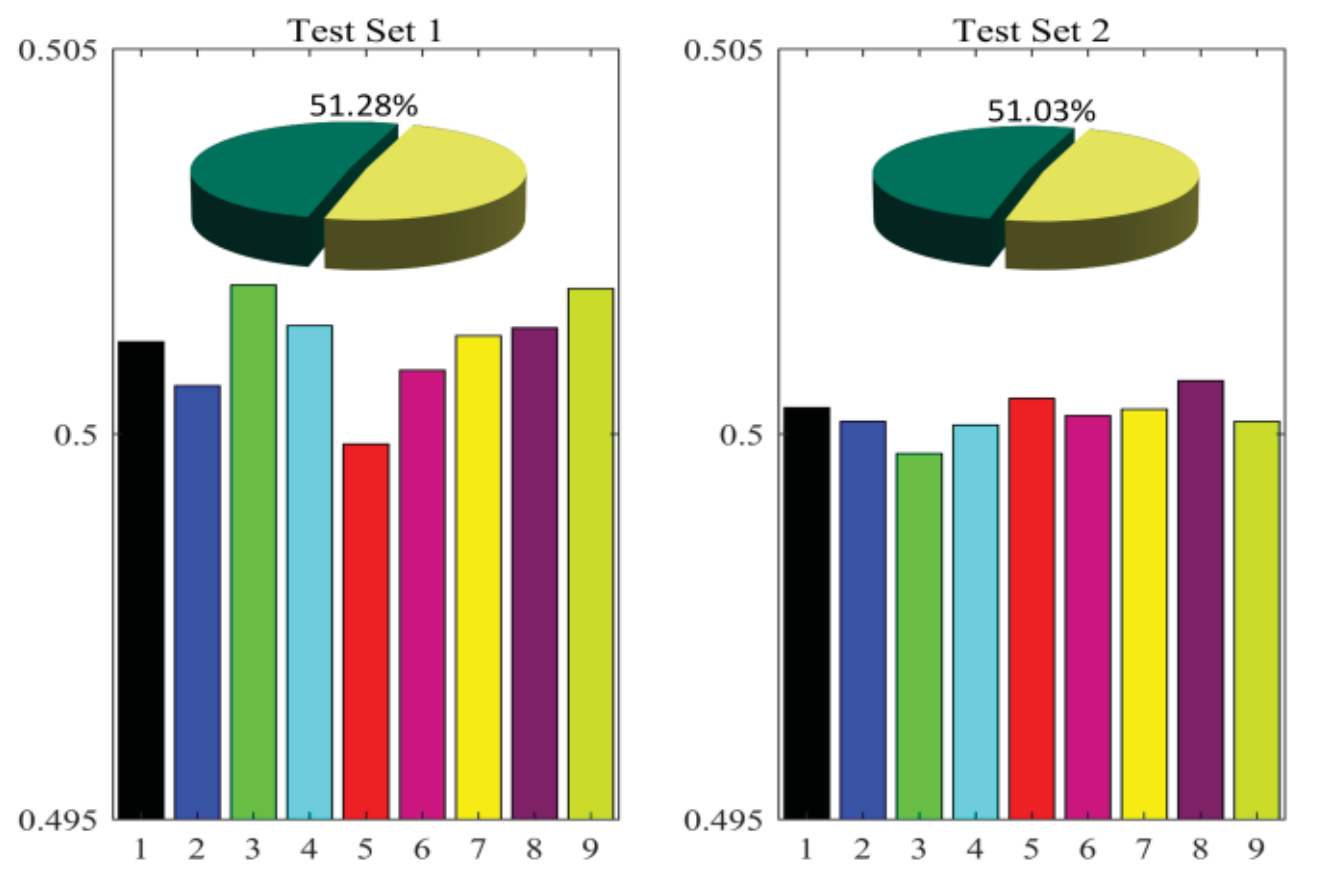

**Figure 2.** Pseudo-randomness of $\pi'$ suppressed via machine learning. Our data demonstrate that the neural network learned from the training dataset performed well in 18 subgroups from two testing datasets respectively. Statistically, the mean of the correct prediction rates goes over the theoretical mean 0.5 with the confidence of 99.9% and 99% for the first and second testing datasets respectively.

For the correct overall prediction rates of 51.28% and 51.03% for the two training datasets respectively, what is the chance that the network achieved the positive performance by randomly guessing? According to the central limit theorem, the random fluctuation in the prediction performance should obey the Gaussian distribution with mean 0.5 and deviation $\sigma = 0.025$. Since $0.5128 > 0.5 + 5\sigma$, $0.5103 > 0.5 + 3\sigma$, we can reject the hypothesis of randomly guessing with the confidence levels of $5\sigma$ and $3\sigma$ for the two testing datasets respectively. Furthermore, according to the student distribution, the one-sided lower confidence limit (LCL) can be calculated by

$$\text{LCL} = \overline{X} - t_{\alpha,n-1}\frac{S_n}{\sqrt{n}}, \tag{2}$$

This means that there is a $100(1-\alpha)\%$ chance for the true mean being over LCL. For the two testing datasets, the confidence probabilities 99.9% and 99% respectively. These arguments show that Eq. (1) is invalid in the case of the sequence $\pi'$. The occurrences of 0 and 1 influence each other in a subtle but definite way.

At the same time, the results from the testing datasets T1 and T2 suggest that the correlation between the knowledge learned by the network from the training dataset is slightly weaker in predicting digits in T2 than T1. We consider this discrepancy reasonable because the distance from the training dataset is longer to T2 than T1 in terms of the number of digits between the training and testing datasets

To make sure that the ability of learning from pseudo-randomness with such a simplistic network is real, we repeated our analysis on two other constants $e$ and $\sqrt{2}$. The protocols were made consistent to that for the investigation on $\pi$. The training and testing datasets were made by grabbing the 1st to 40007th and 100000th to 1000006th digits respectively from each of the digital series. As far as the learning results are concerned, In the case of $e$, the correct prediction rates for 9 subgroups were 0.50079, 0.50075, 0.50003, 0.50130, 0.50164, 0.50255, 0.50163, 0.50098 and 0.50086 respectively; and in the case of $\sqrt{2}$, the measures were 0.50016, 0.50097, 0.50096, 0.49996, 0.49938, 0.50102, 0.50163, 0.50162 and 0.50194 respectively. With confidence probability 99%, the true means of both are over LCL (>0.5). Furthermore, the performance in the training datasets of e and $\sqrt{2}$ are 0.5092 ($>0.5+3\sigma$), 0.5132 ($>0.5+5\sigma$) respectively.

### *B. Dealing with Multiple Pseudo-Random Sequences*

The pseudo-random number generator (PRNG) generates random numbers that are practically indistinguishable from ideal random numbers, and widely used in various Monte Carlo simulation studies. Usually, a sequence of pseudo-random numbers is initiated by a "seed" inside the PRNG. Once the seed is reset, the number sequence is thoroughly altered. After we have revealed correlative patterns in a single pseudo-random sequence such as the sequence $\pi'$, we are motivated to evaluate the potential of the neural network in suppressing pseudo-randomness of the Mersenne Twister (MT) associated with multiple initial seeds [19], since MT is a most reliable and widely used PRNG.

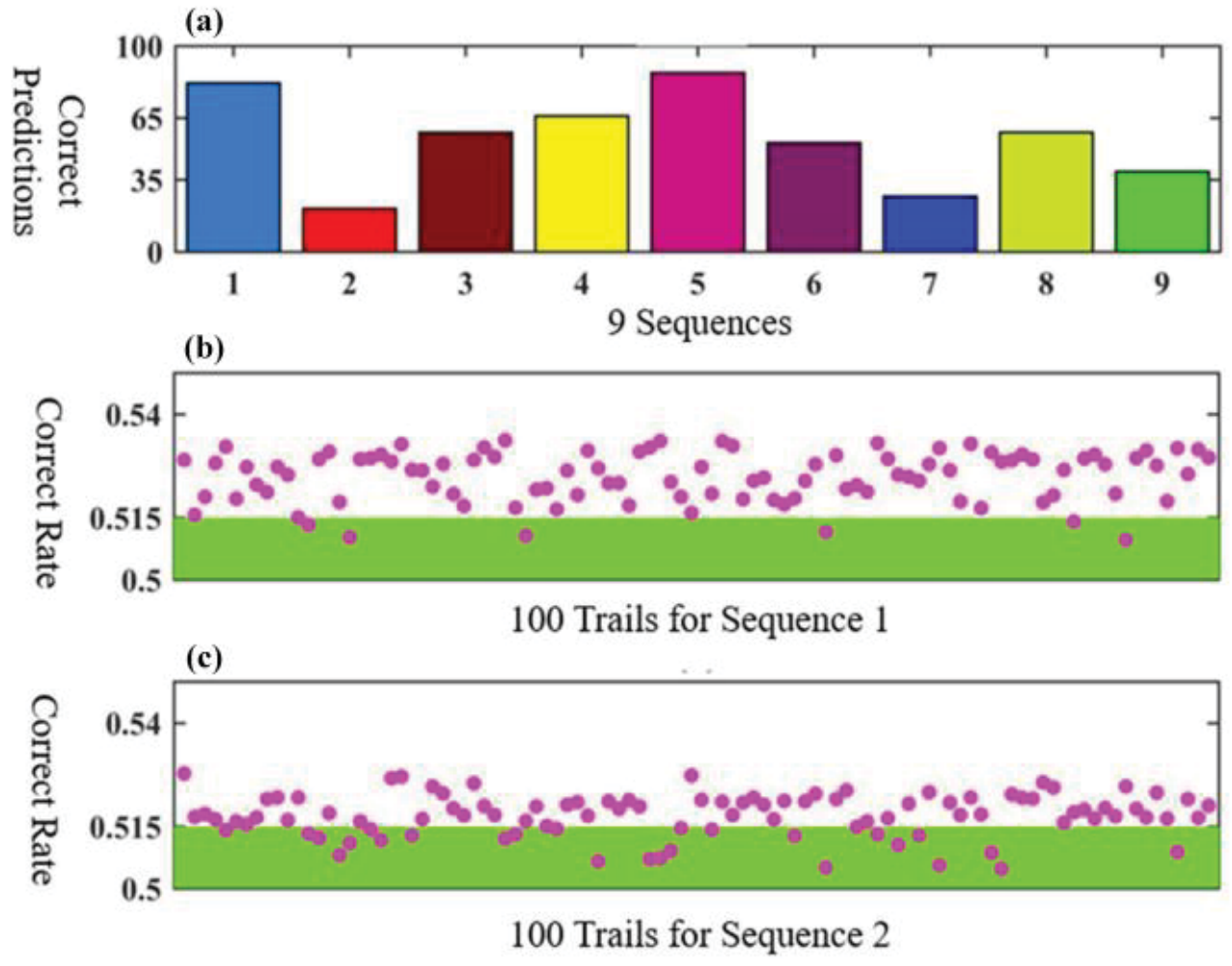

**Figure 3.** Performance fluctuation in predicting random members from MT. (a) For 10 random sequences, the number of successful predictions are 82, 21, 58, 66, 87, 53, 27, 58 and 39 out of 100 respectively, and (b) and (c) the percentages of correct predictions are 95 and 77 out of 100 for the first two sequences respectively.

In MatLab, the range of MT is [0,1]. We used a threshold 0.5 to generate a random 0-1 sequence. Consistent to our first study, the same network structure and training protocol were used to predict the seventh number from its precedent consecutive six numbers, with the following training parameters: momentum factor=0.95, learning rate=0.05, maximum epoch=100, and minimum gradient=1e-10. We arbitrarily generated 9 random sequences and binarized them. For each sequence, the training dataset consisted of 10,000 instances from the 1st to 10,006th numbers. The testing dataset consisted of only one instance from the 10,001-10,007 numbers in the corresponding sequence. In this way, the trained network will be applied to the closest possible location for a maximized correlation. Then, we repeated such a process by 100 times and counted the total number of successful predictions for each of the 9 sequences. As shown in Figure 3, the number of correct predictions were found to be 82, 21, 58, 66, 87, 53, 27, 58, and 39 out of 100 respectively.

Why do the prediction rates differ significantly? We investigated into the underlying reason in terms of the network performance for the training dataset in 100 independent trials (initialized by fresh randomization) of the first two sequences respectively. During each training session, we recorded the prediction performance as shown in Figure 3(b) and 3(c) for the two sequences respectively. Since 95 and 77 of 100 exceeded the 51.5% for the two sequences respectively, we infer that the training did enable the network to detect a correlation in each training dataset. Then, why did the prediction results for the testing dataset were polarized (82 and 21 out of 100 for the two sequences respectively)?

Let us assume that the input pattern consisting of 6 consecutive random numbers "abcdef" and the output is the next binary number 0 or 1. If Probability (1 | abcdef) ≠ Probability (0 | abcdef) ≠ 0.5, then a well-trained network should favor the number with higher probability, implied when the loss function was minimized to reflect the statistical bias. In the first sequence, the number of occurrences of the combined 7-number pattern "011000-1" was counted as 88, while the number of occurrences of the complementary pattern "011000-0" was counted as 78. Whenever the instance in a testing dataset is 011000-1, it should not be surprising to achieve a successful prediction, as was the case for the first sequence and reaching 82 out of 100 trials (again, in each of which the network was independently trained). In the second random sequence, there were 78 7-number patterns "111111-0" and 65 complementary patterns "111111-1". Unfortunately, the instance in the corresponding testing dataset was "111111-1" (happens to be against the favorable pattern of the network, which is "111111-0"), which explains such a low rate of successful predictions. Note that these results, either higher or lower than 50 of 100, are still in agreement with our explanation. Now, assuming both training and testing datasets are huge. According to the law of large numbers, the proportion of any string in both sets will be close to the corresponding conditional probability distribution. Therefore, the optimization of the neural network ensures not only an effective learning of whatsoever statistical relationships in training data but also a successful generalized into testing data as long as the former and latter are from the same distributions.

The above analysis supports the intrinsic capability of the neural network in learning probabilistic knowledge and inferring statistically. The neural network can effectively learn whatsoever probabilistic associations in a training dataset. Then, the neural network can use the knowledge or rules learned from the training data in predicting outcomes in testing data. If the testing data share the same statistical properties with less than the maximum entropy, then the neural network will be able to do prediction or association as effectively as the information content of the training data allows. Even in rather challenging cases such as pseudo-random number sequences from transcendental numbers and best random number generators, very simplistic feedforward neural networks can do a quite impressive job as exemplified above.

### C. *Testing for Number Normality*

A normal number refers to a number $S$ whose sequence of consecutive digits satisfying the limit [20]:

$$\lim_{n\to oo}\frac{N_s(w,n)}{n}=\frac{1}{b^{|w|}} \quad (3)$$

where $w$ is any finite string of digits, $N_s(w,n)$ is the total number of times that $w$ occurs in the first $n$ digits of the number, and $b$ is the base of $S$. The definition of the normality reflects the nature of randomness [11]. Nevertheless, how to determine if a number is normal is still an open problem. In particular, it is not known if $\pi$, or $\pi'$ (in a binary disguise), is normal or not [21-22], although the first 30 million digits of $\pi$ are quite uniform [18]. With $S=\pi'$, if $\pi'$ is a normal number, then any string $w$ with 7 digits will satisfy $\lim_{n\to oo}\frac{N_s(\omega,n)}{n}=\frac{1}{2^7}$. In particular, $\lim_{n\to oo}\frac{N_s(0011001,n)}{n}=\frac{1}{2^7}$ and $\lim_{n\to oo}\frac{N_s(0011000,n)}{n}=\frac{1}{2^7}$, which means whether the seventh digit takes 0 or not is statistically independent with the first six digits "001100", and any machine learning attempt would fail.

We would like to present an alternative definition to characterize the normality. As we know, there are in total $b^{|\tau|}$ possible strings of a finite length $\tau$. For prediction of the next digit, we define a "a group of mutually-exclusive strings" $G_{\tau'}$ as { $\tau'$-0, $\tau'$-1, $\tau'$-2, $\tau'$-3…… $\tau'$-(b-1)}, where $\tau'$ is a finite string with $|\tau|$-1 digits. Then, there are $b^{|\tau|-1}$ mutually exclusive groups. Given the first n digits of S, we can record the number of occurrences of each element in $G_{\tau'}$, and denote that occurs the most often as $N_s(\tau,n)^{(\tau')}_{\max}$. We propose to use the following formula

$$\sum_{\tau'}N_s(\tau,n)^{(\tau')}_{\max}\le\frac{1}{b}+\frac{5\sqrt{b-1}}{b\sqrt{n}} \quad (4)$$

to test the normality.

Clearly, Eq. (3) can be deduced from Eq. (4). In the perspective of prediction, with a given $\tau'$ the predictor can only have one output, and the best performance for the ideal predictor must be $\frac{\sum_{\omega'}N_s(\omega,n)^{(w')}_{\max}}{n}$. We define $Y=\frac{\sum_{k=1}^{n}X_k}{n}$, where the random variable $X_k=1$ when the prediction is correct, and otherwise $X_k=0$. If the prediction by the predictor is completely by guessing, then Y has the mean of $\frac{1}{b}$ and deviation of $\frac{\sqrt{b-1}}{b\sqrt{n}}$. If Eq. (4) does not hold, we have a $5\sigma$ level confidence to claim that the predictor does not work purely by chance. In this way, Eq. (4) establishes a more straightforward connection between normality and pseudo-randomness than Eq. (3).

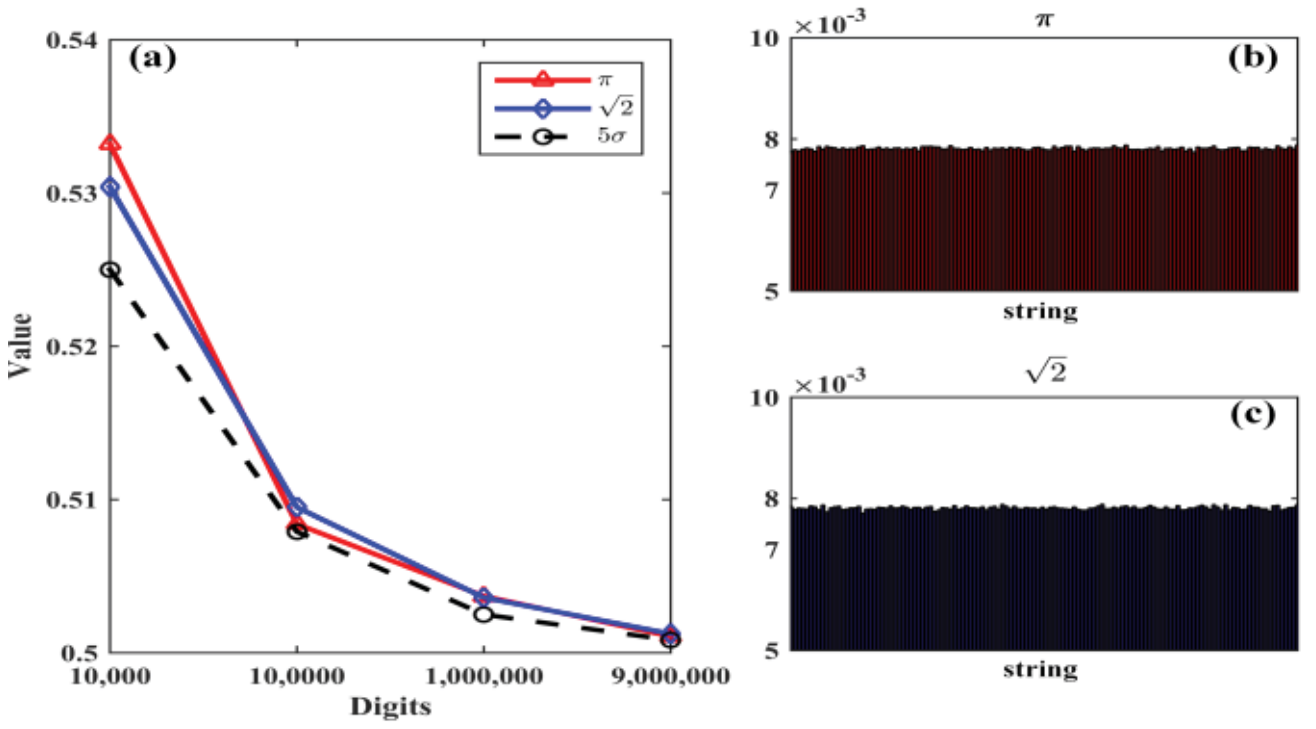

**Figure 4.** In the 0-1 sequence of $\pi$ and $\sqrt{2}$, with $|\tau|$=7 and $b$=2 the evaluation of Eqs. (3) and (4) indicate that the two 0-1 sequences seem not perfectly normal.

Motivated by Eq. (4), we binarized the sequences of $\pi$ and $\sqrt{2}$ respectively. Then, with $|\tau|$=7 and b=2 we calculated $\sum_{\tau'}N_s(\tau,n)^{(\tau')}_{\max}$ in each case for n=$10^4$, $10^5$, $10^6$ and 0.9*$10^7$ respectively, as shown in Figure 4(a). The black broken line represents the $5\sigma$ bound. In addition, for $n$=9,000,000 we individually counted $N_s(\tau,n)$ and calculated $\frac{N_s(\tau,n)}{n}$ for every finite string and made the bar charts. It is observed in Figure 4(b) and (c) that the frequencies of these strings are pretty much at the level of $\frac{1}{128}$. Moreover, the frequencies of 1 or 0 in the binarized $\pi$ and $\sqrt{2}$ sequences tend to converge to 0.5 when n increases. These facts suggest that the 0-1 sequences of $\pi$ and $\sqrt{2}$ are in agreement with the current definition of normality. However, the data from $\pi$ and $\sqrt{2}$ in Figure 4(a) are yet beyond the black broken line, violating Eq. (4). Therefore, we argue that the first $10^7$ digits of either binary sequence are not perfect normal by our $5\sigma$ ruler Eq. (4). By the way, the imperfect normality of $\pi$ and $\sqrt{2}$ are not un-imaginable, since either can be put into a quite regular series expansion. Basically, the imperfect normality of $\pi'$ not only accounts for the $5\sigma$ performance in our first numerical experiment but also sets an upper limit of the network performance.

## III. Discussions and Conclusion

Assuming that A is a presetting and B is an emerged event, $P(B|A)=1$ denotes a definite association between A and B, while $P(B|A)=P(A)$ means a statically independence of B from A. Our study has highlighted a key utility of the neural network in the cases of $P(B|A)\simeq P(B)$ to detect any nuance of correlation among data and realize any hidden information for a purpose such as prediction. While the complexity of neural network is more than conventional statistical testing in terms of detecting any difference between P(B|A) and P(B), the asymptotical tight bound of both methods are $\Theta(n)$, which means their computational complexities are comparable. Moreover, a trained well-designed network itself is an explicit low-dimensional expression of the underlying relationship from inputs to outputs. It is surprising that the neural network, even just a simplistic feed forward one, could perform this task quite meaningfully, as evidenced above.

A major point we want to make is that the use of a neural network in extremely uncertain environments such as with pseudo-randomness can be interesting and promising. Our numerical simulation suggests a non-trivial extension of what machine learning can do. More powerful architectures of neural networks, such as GAN and RNN, could bring more insight into pseudo-randomness. It is worthwhile further investigating the potential of the neural network in learning on a low-dimensional manifold, not only in statistical terms but also beyond (such as identifying the structure of a PRNG). The practical implication of research along this line is enormous, such as for detection of rather weak signals in a strong noisy environment or cope with chaotic dynamics to a certain degree (making money in the stock market).

Modern theories with fundamentally uncertain properties play a pivotal role in depicting and transforming the world, such as the unpredictability of particles on a quantum scale and chaotic behaviors due to classic nonlinearities. Initially, Schrödinger formulated the electron's wave function as the charge density across the field, and then Born reinterpreted it as the electron's probability density. Up to now, the probabilistic viewpoint remains the mainstream in the field of quantum mechanism. Through most intensive debates against strongest opponents such as Einstein, one of his famous quote is "*God does not play dice with the universe*", the probabilistic theory of quantum mechanism has been well established as rigorously tested, self-consistent and complete. A question we would like to ask is if the randomness exhibited in quantum mechanics is genuinely random or pseudo-random? The question can be answered with a statistical hypothesis test, which can be designed in various ways and seems now addressable physically and computationally using the neural network approach.

Most trustable random numbers are generated according to quantum mechanics, which have applications especially in cryptography. Randomness is in principle pseudo-randomness from the perspective of classical physics. The generation of quantum randomness is impossible with classical means. Quantum random number generators (QRNGs) are believed to give genuine randomness, and currently categorized into three classes [23]: QRNGs built on trusted and calibrated devices, QRNGs with verifiable randomness without trusting the implementation, and those between the first two classes. Given the application-oriented motivation, all these QRNGs were designed to seek a balance between efficiency and randomness eventually optimizing cost-effectiveness. On the other hand, it is intellectually interesting to test for any hint that quantum randomness might be a kind of pseudo-randomness. A QRNG is needed for such a test, which should be both statistically rigorous and technically manageable.

Figure 5 is a new QRNG design (to our knowledge), in which a laser experiment for double-slit diffraction is utilized so that there is no any classic component (such as a beam splitter) between a coming photon and its detector). After the wave interference, any variations in the laser light source and double-slit screen details are "averaged" out. Then, both the valleys around the main interference lobe offer ideal separating zone defining two symmetric parts (this partition can be made even more distinct if we only detect the so-called "pile-up" effect). Then, hits on the top detecting region are interpreted as "1"s, while hits on the bottom detecting regions are recorded as "0"s in the photon-counting mode. A coincidence detection circuit similar to that used for positron emission tomography can reject simultaneous arrivals of photons. Furthermore, the detecting regions can be pixelated to increase statistical power. This experimental design is based on mature technologies, keeping the quantum nature as perfectly as feasible. With this QRNG or alike, we can critically examine any difference between inherent and pseudo random numbers using the above-described neural network approach; in other words, we can make sure if the realization of a wave function is ideally random without any subtle bias (maximum entropy) or there is a definite deviation (even very tiny) from symmetry. If the latter case could be validated up to a 5σ confidence level, then we know that what God is playing is a pseudo-dice.

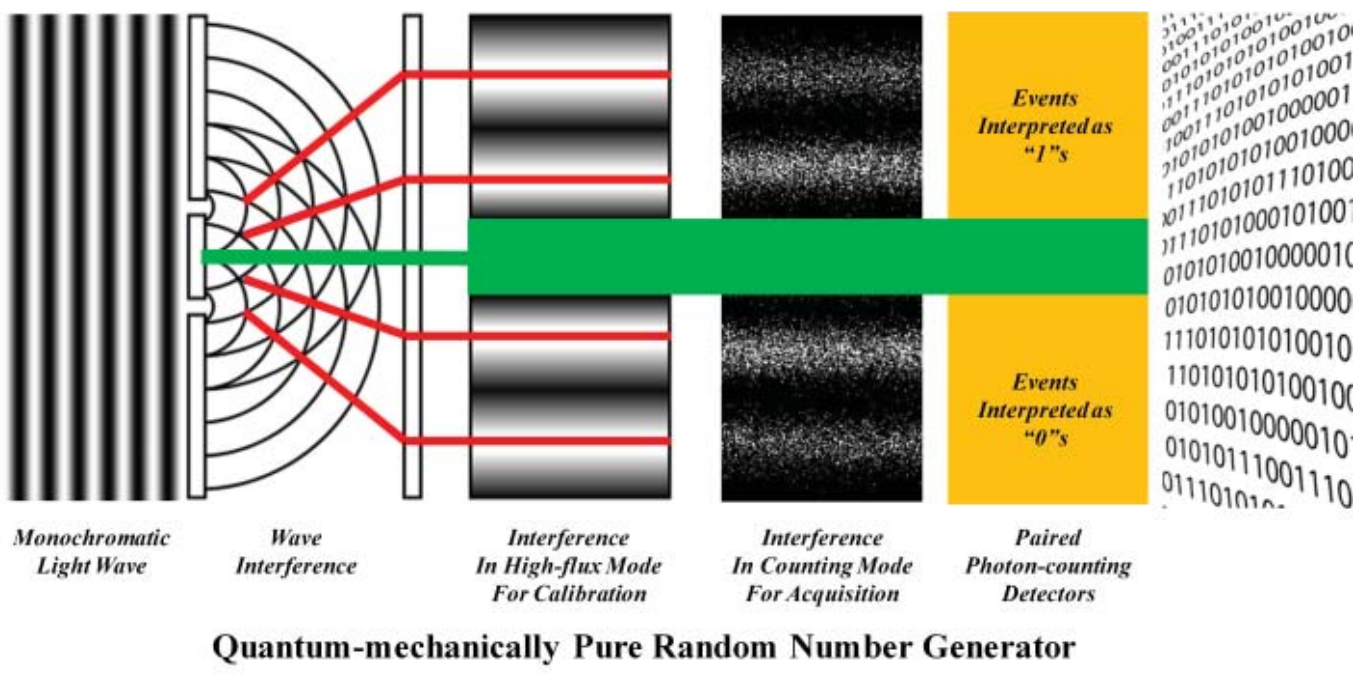

**Figure 5.** *Conceptual design of a physical 0-1random number generator.* This design should be significantly refined for accuracy and efficiency.